\documentclass[runningheads]{llncs}
\usepackage[T1]{fontenc}
\usepackage{graphicx}
\usepackage{placeins}
\usepackage{amsmath}
\usepackage{cite}
\usepackage{amsfonts,amssymb}
\usepackage{tabularx}

\begin{document}
\title{AMSA-UNet: An Asymmetric Multiple Scales U-net Based on Self-attention for Deblurring}
\titlerunning{AMSA-UNet for Deblurring}
%
\author{Yingying Wang\inst{1}}
\authorrunning{Y.Y. Wang}
%
\institute{Shenyang Aerospace University, Shenyang LN 110136,CHINA\\
\email{ckertwong@gmail.com}}
\maketitle              
\begin{abstract}
The traditional single-scale U-Net often leads to the loss of spatial information during deblurring, which affects the deblurring accuracy. Additionally, due to the convolutional method's limitation in capturing long-range dependencies, the quality of the recovered image is degraded. To address the above problems, an asymmetric multiple scales U-net based on self-attention(AMSA-UNet) is proposed to improve the performance of deblurring methods in accuracy and computational complexity. By introducing a multiple-scales U shape architecture, the network can focus on blurry regions at the global level and better recover image details at the local level. In order to overcome the limitations of traditional convolutional methods in capturing the long-range dependencies of information, a self-attention mechanism is introduced into the decoder part of the backbone network, which significantly increases the model's receptive field, enabling it to pay more attention to the semantic information of the image, thereby producing more accurate and visually pleasing deblurred images. What’s more, a frequency domain-based computation method was introduced to reduces the computation amount. The experimental results demonstrate that the proposed method exhibits significant improvements in both accuracy and speed compared to eight excellent methods.

\keywords{Deblur \and U-Net \and Self-attention mechanism \and Fast Fourier Transform.}
\end{abstract}
\section{Introduction}
Earlier deblurring methods primarily focused on non-blind deblurring, recovering images with known blur kernels. Pan et al.~\cite{ref_article1} accurately computed blur kernels using the dark channel's sparsity in blurry images to restore sharp images. However, these traditional approaches struggle with spatially varying blur and are often time-consuming.

With advancements in deep learning, CNN-based non-blind deblurring methods have gained prominence. Include CNN-based blur kernel estimation and deconvolution networks using estimated kernels. Sun et al.~\cite{ref_lncs1} used CNNs to enhance motion smoothness in blur kernels, while Chakrabarti et al.~\cite{ref_lncs2} modeled blur kernel coefficients for accurate deconvolution. Combining traditional and deep learning methods has improved deblurring, yet challenges remain in handling occlusions, depth variations, and noise sensitivity in blur kernel estimation, limiting their effectiveness in complex scenarios.

To overcome the limitations of non-blind deblurring, end-to-end methods using convolutional neural networks have been developed. These methods directly map blurred images to sharp images, avoiding reliance on blur kernel estimation. Nah et al.~\cite{ref_lncs3} employed a multi-scale coarse-to-fine approach for effective deblurring in dynamic scenes. The DeepDeblur model by Mei et al.~\cite{ref_article2} is noted for removing text blur in document images using an end-to-end approach. However, these methods, due to complex network structures, require extensive training time. DeblurGAN by Kupyn et al.~\cite{ref_lncs4} incorporates generative adversarial networks (GANs) for deblurring, achieving good results through a game between generator and discriminator. Despite the progress made by these methods, issues such as stability of training and computational size limit their application.

In recent years, researchers have introduced the fully convolutional neural net-work U-Net to enhance the focus on semantic information in images. U-Net improves model accuracy to extent, but its single network structure leads to a significant number of redundant calculations during processing, resulting in spatial information loss. To address the spatial information loss problem of U-Net, Cho et al.~\cite{ref_lncs5} proposed a coarse-to-fine strategy to better preserve the feature information of im-ages. However, the efficiency and accuracy of this method still need to be improved in capturing long-range dependency relationships.

With the development of the transformer, the self-attention mechanism has also been introduced into the field of image deblurring. Restormer, an efficient model based on Transformer proposed by Zamir et al.~\cite{ref_lncs6}, has shown promising results in high-quality image restoration tasks. However, the traditional Transformer model struggles to effectively capture the relationships among local pixels in pixel-level data processing, leading to unnecessary computational overhead.

Therefore, an asymmetric multiple scales U-net based on self-attention is proposed. The overall image deblurring accuracy is improved by introducing an attention mechanism and spatial information is preserved to a greater extent by the multi-scale architecture. 
\section{Related Work}
\subsection{Multi-scale deep convolutional network deblurring methods}
Deep Multi-scale Convolutional Neural Network, serves as a pioneering exploration of end-to-end learning strategies, successfully utilizing deep convolutional networks to process images at different scales. This network design exquisitely adopts a coarse-to-fine strategy to progressively restore image clarity. In this process, the inputs of each layer of the network cleverly connect the outputs of the coarse-scale network with the inputs of the fine-scale network, realizing the smooth transmission of information from coarse to fine. The processing of each layer of the network rigorously following Eq. (1).
\begin{equation}
B=\frac{1}{M}\sum_{i=0}^{M-1}S[i]
\end{equation}
where $\text{B}$ denote a blurry image and $\text{S[i]}$ represent the $i_{\text{th}}$ sharp image, $\text{M}$ is the number of clear images. 

Inspired by DeepDeblur~\cite{ref_article2}, researchers further gained insights into the non-independence among the parameters of network layers at different scales, as well as the differential effects of image blurring on images at different scales~\cite{ref_lncs7}. Thus, a Parameter Selective Sharing and Nested Skip Connections (PSS-NSC) structure has been proposed. This structure effectively solves the gradient vanishing problem, enabling better preservation and recovery of details in the image reconstruction process. Its reconstruction process following Eq. (2).
\begin{equation}
\hat{S}=F_{\left(a_{n}^{P},a^{P}\right)}^{P}(B_{n};\left(\hat{S}_{n+1}\right)^{\uparrow})+B_{n}
\end{equation}
where $F_{\left(\alpha_{n}^{P},\alpha^{P}\right)}^{P}$ denotes the sub-network of the $n_{th}$ layer, the parameters are com-posed of the parameters of this layer $\alpha_{n}^{P}$  and the shared parameters ${\alpha}^{p}$ , $B_{n}$ is the blurry image of the $n_{th}$ layer, denotes the image processed by the 
$n_{th}$ layer of the network, and the symbols $\uparrow$ denote the upsampling operations.

Through the above improvements, the two networks provide new ideas and directions for future research in multi-scale processing.
\subsection{Transformer Deblurring Method}
Transformer exhibits significant advantages in capturing long-range dependencies in information due to its inimitable global attention mechanism~\cite{ref_lncs8}. Transformer is able to access a wider receptive field compared to traditional convolutional structures, enabling the model to focus on contextual information, and thus generating more faithful image outputs. As a result, researchers have introduced the self-attention mechanism into image deblurring tasks.

However, simply applying self-attention mechanisms to image deblurring tasks leads to a significant increase in computational cost. To solve this problem, a method for computing scaled dot-product attention in the feature depth domain has been introduced, enabling efficient extraction of feature information from different feature channels. In addition, a multi-head attention mechanism in conjunction with a multi-scale hierarchical module were proposed, which not only enhances the representation capability of the model, but also effectively reduces the computational cost. The multi-head attention mechanism processing following Eq. (3).
\begin{equation}
    \begin{gathered}
    \hat{X}=W_{p}\text{Attention}(\hat{Q},\hat{K},\hat{V})+X \\
    \text{Attention}(\hat{Q},\hat{K},\hat{V})=\hat{V}\cdot \text{Softmax}\left(\frac{\hat{Q}\cdot\hat{K}}{\alpha}\right)
    \end{gathered}
\end{equation}
where $Q, K, V$ and $\hat{X}$ denote the input and output feature maps, respectively, $\hat{Q}$ , $\hat{K}$ and $\hat{V}$ are new matrices obtained by reshaping the size of the matrix from the original one of size $\mathbb{R}^{\hat{H}_{\times}\hat{W}_{\times}\hat{C}}$ , and $\text{a}$ is a learnable scaling parameter used to control the size of the dot product of $\hat{K}$ and $\hat{V}$ in the $\text{Softmax}$ function.

Inspired by previous work, a locally-enhanced window attention mechanism module (Locally-enhanced Window block, LeWin block) was proposed~\cite{ref_lncs9}. The module contains two parts, Window-based Multi-head Self-Attention (W-MSA) and Locally-enhanced Feed-Forward Network (LeFF). The W-MSA module processing as following Eq. (4).
\begin{equation}
    \begin{gathered}
    Y_{k}^{i}=Attention(X^{i}W_{k}^{Q},X^{i}W_{k}^{K},X^{i}W_{k}^{F}),i=1,...\frac{HW}{M^{2}}\\
    Attention(Q,K,V)=Soft\max(\frac{Q\cdot K}{C/k})\\
    \hat{X}_{k}=\{Y_{k}^{i}\},i=1,...,\frac{HW}{M^{2}}
    \end{gathered}
\end{equation}
where $X^{i}$ denotes the localized region of the input feature $\mathbf{X}\in\mathbb{R}^{C\times H\times W}$ with $\text{C}$ , $\text{H}$ and $\text{W}$ being its channel number as well as its height and width, respectively, and $Y_{k}^{i}$ is the output of the $k_{th}$ header to the $i_{th}$ region, combining all $Y_{k}^{i}$ together con-stitutes the entire output of the $k_{th}$ header $\hat{X}_{k}$ . The output of the W-MSA module is processed by LeFF and superimposed on the original input to obtain the overall output of the LeWin block. The module makes full use of the advantage of the self-attention mechanism in capturing long-range dependencies, which enables the model to prioritize semantic information in blurry regions and enhances its capability to process image details.

Although these methods have made some progress in terms of accuracy, they still need to go through complex matrix multiplication operations, and their complexity is still $O(n^{2})$ . Therefore, how to further reduce the computational complexity while maintaining the performance is still an important direction for future re-search. Possible solutions include optimizing the computation of the attention mechanism, adopting more efficient feature extraction methods, or introducing new model architectures.
\subsection{Fourier Transform}
The convolution theorem is an important principle in signal processing that shows the equivalence of convolution operations in the time domain to multiplication operations in the frequency domain. This property is also applicable in image processing, allowing complex convolution calculations in the time domain to be simplified by frequency domain operations. Specifically, by utilizing Fast Fourier Transform (FFT) and Inverse Fast Fourier Transform (IFFT), convolution operations that would otherwise require a time complexity of $O(n^{2})$ can be accomplished in a time complexity of $O(n\log n)$ . The FFT's mathematical formulation is presented in Eq. (5).
\begin{equation}
f*g=F^{-1}\{F\{f\}•F\{g\}\}
\end{equation}
the convolution operation for $\text{f}$ and $\text{g}$ is equivalent to performing a Fast Fourier Transform $F\{f\}$ , $F\{g\}$ and then transferring it to the frequency domain for the multiplication operation, after which it is converted back to the time domain using the Fast Fourier Inverse Transform.

In the field of image deblurring, many studies have fully utilized this property of the convolution theorem~\cite{ref_lncs10,ref_lncs11,ref_article3}. By utilizing the convolution theorem and the fast Fourier transform, the researchers were able to reduce the computational complexity of image processing tasks while maintaining performance. This approach not only improves processing speed, but also helps build more efficient and lightweight deep learning models, providing more possibilities for practical applications.
\section{Proposed Method}
\begin{figure}
\includegraphics[width=\textwidth]{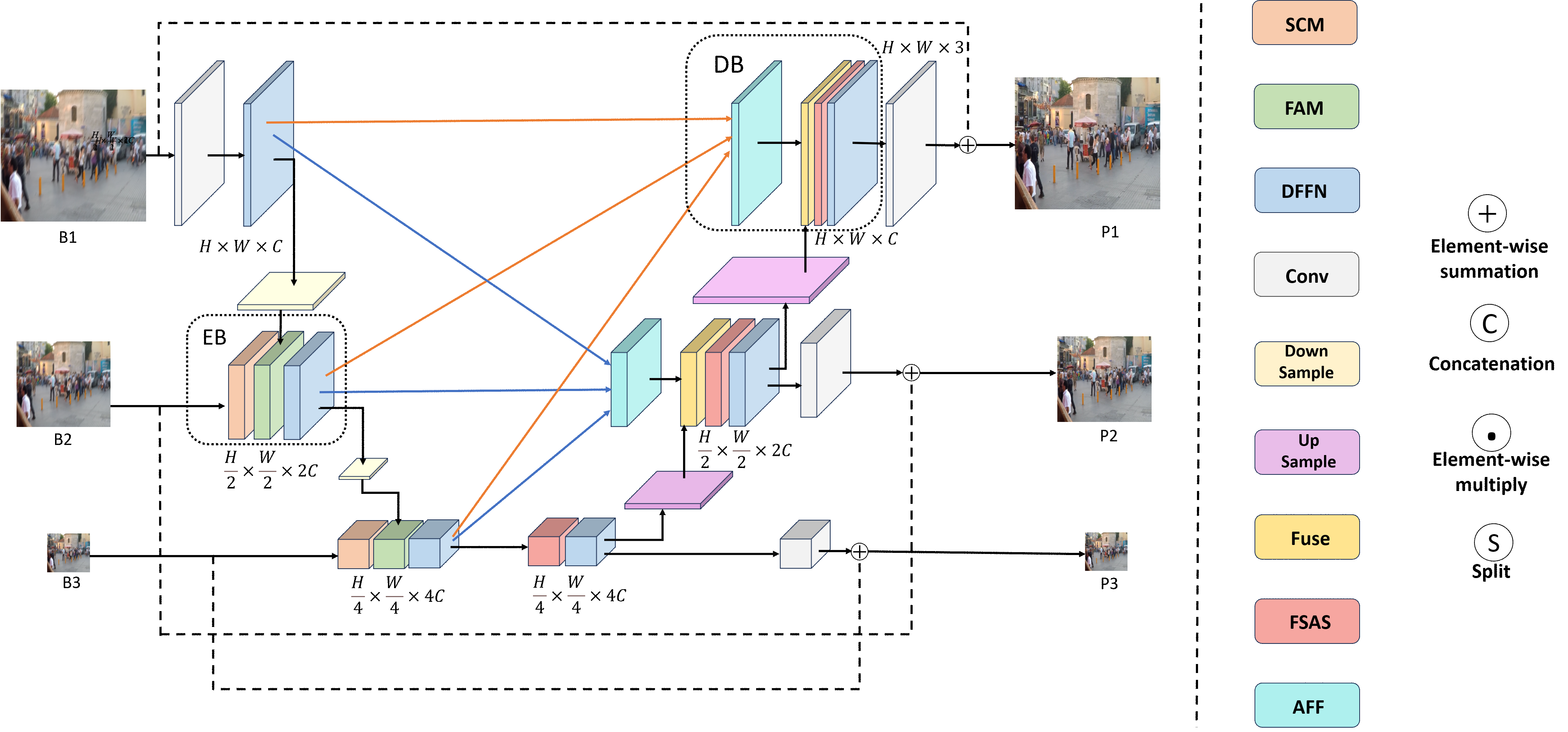}
\caption{Overall Network Architecture} \label{fig1}
\end{figure}

In this paper, an asymmetric multiple scales U-net based on self-attention(AMSA-UNet) is proposed which combines the multiple-input-multiple-output network architecture with the transformer module, in order to solve the loss of image spatial features induced by the single-scale U-Net network. This method enhances the model's receptive field by introducing self-attention into the decoder module. At the same time, the Fourier transform is utilized to improve the computational ability of the model and reduce its computational complexity. Additionally, the multi-scale feature fusion module in the traditional Multiple Inputs and Multiple Outputs-U-Net (MIMO-UNet)~\cite{ref_lncs12} architecture is retained to further enhance the model's ability to learn multi-scale image information and generalization ability. The overall framework of the network is shown in Fig.~\ref{fig1}
\subsection{Encoder block}
In order to reduce the feature information lost in the downsampling process of the model and to effectively resolve the ambiguities at different scales, the Encoder adopts a multi-scale input strategy, the downsampling of the upper layer inputs is combined with the upper outputs as the inputs to the current layer. 

In encoder block, a shallow convolutional module (SCM) is used for feature ex-traction from the downsampled image, where the SCM is a pathway consisting of two sets of 3×3 and 1×1 convolutional layers combined to process the original input and output the result after connecting it to the original input, the structure of the SCM is shown in Fig. 2. The output of the SCM module and the output of the upper layer encoding block are fused by the feature attention module (FAM), which can emphasize the effective information of the previous scale or suppress the unnecessary information and learn the spatial information from the output of the SCM, so that the inputs of the encoder block of each layer contain the most effective multi-scale information. Therefore the input of the FAM module will be taken as the input of the encoder block of this layer. The specific FAM structure is shown in Fig. 2

\begin{figure}
\includegraphics[width=\textwidth]{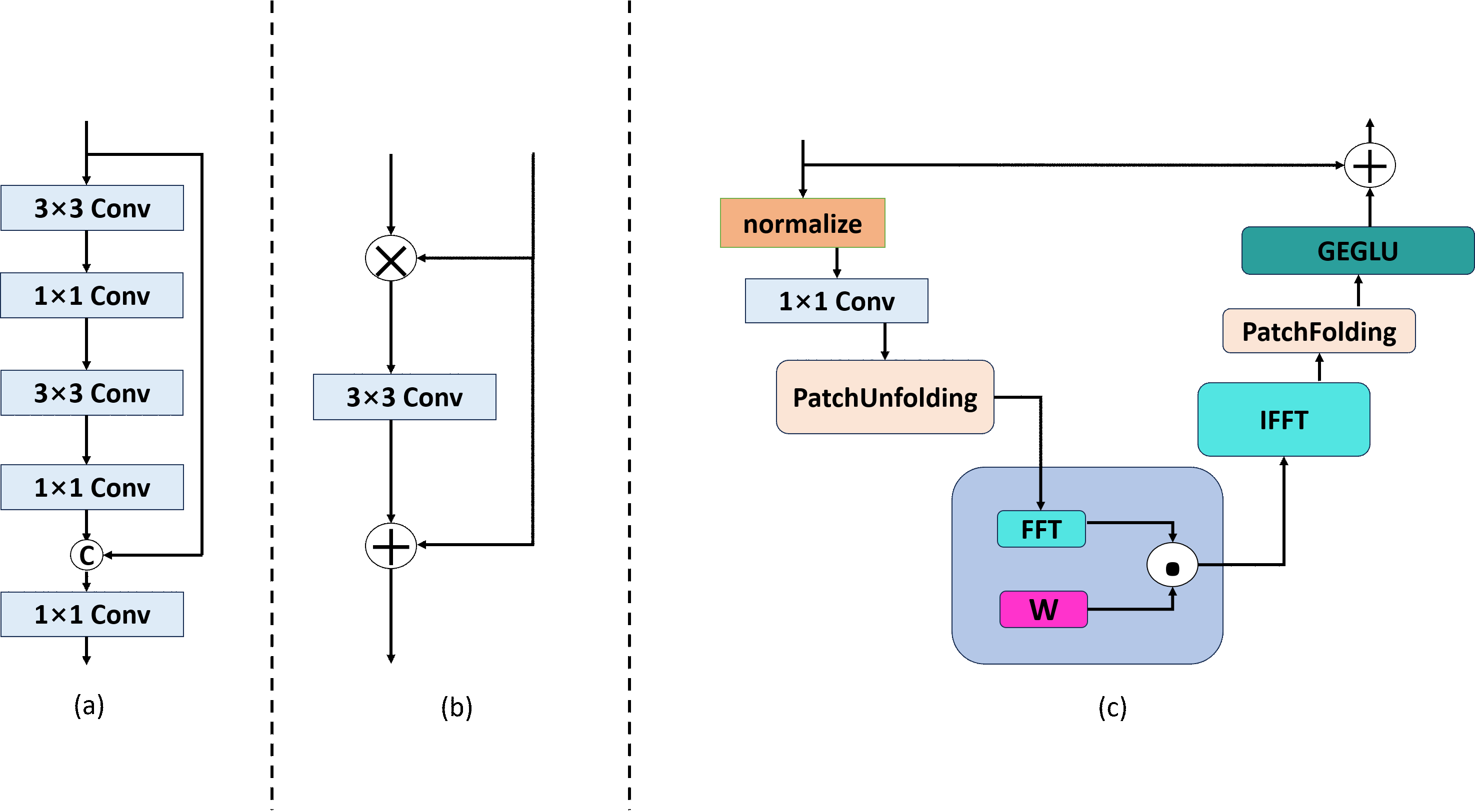}
\caption{The Structure of encoder block. (a) SCM, (b) FAM, (c) DFFN.} \label{fig2}
\end{figure}

Since not all frequency information is valid for the deblurring task, a frequency domain-based feedforward neural network (DFFN) is introduced as the main component of the encoder block to determine which frequency information should be retained for better reconstruction of sharp images. In order to discriminate the valid frequency information, inspired by the compressed image algorithm, a learnable quantization matrix W is introduced in the DFFN, which is learned by the inverse of the compression method to determine which frequency information is valid. In the feedforward channel, the input is converted to frequency domain information by FFT, and then multiplied with the quantization matrix W. The final output is a combination of the input and the channel output, and the output of the DFFN module will be used as part of the final output of the encoder block of the current layer for the next layer of the multi-scale input. The detailed DFFN module structure is shown in Fig.~\ref{fig2}
\subsection{Decoder block}
In the ASMA-UNet, the Decoder consists of three layers of decoder blocks. Each layer of decoder block outputs different-scale results, and the fusion of the lower layer output with the upper layer feature maps as the input to the current layer decoder block can realize the application of intermediate supervision to each decoder block and enhance the accuracy of the model output results.

The input of each layer decoder block is obtained from the output from the lower layer decoder block and the output of the Asymmetric feature fusion (AFF) module through the Fusion Extraction (Fuse) module. The AFF module connects the outputs of the three layers of encoder modules through a 1×1 convolutional layer and a 3×3 convolutional layer to obtain outputs at different scales, so that the information at different scales can be fully utilized. In the network, the first layer and the second layer decoder block will accept the output of the corresponding scale AFF module and the output of the lower layer decoder module to be processed by the feature fusion module Fuse as the input of the decoder module of this layer. In the feature fusion module Fuse, it connects the output of the AFF module and the output of the lower layer decoder module and inputs them into the DFFN module, and then the outputs are divided and summed to obtain the final result after fusion. Unlike the traditional U-Net network, which directly fuses the output of the same layer encoder with the output of the decoder, the existence of the AFF module allows the fusion result to contain more spatial information of the scale, which improves the accuracy of the decoder block. The structure of Fuse and AFF is shown in Fig. 3.

\begin{figure}
\includegraphics[width=\textwidth]{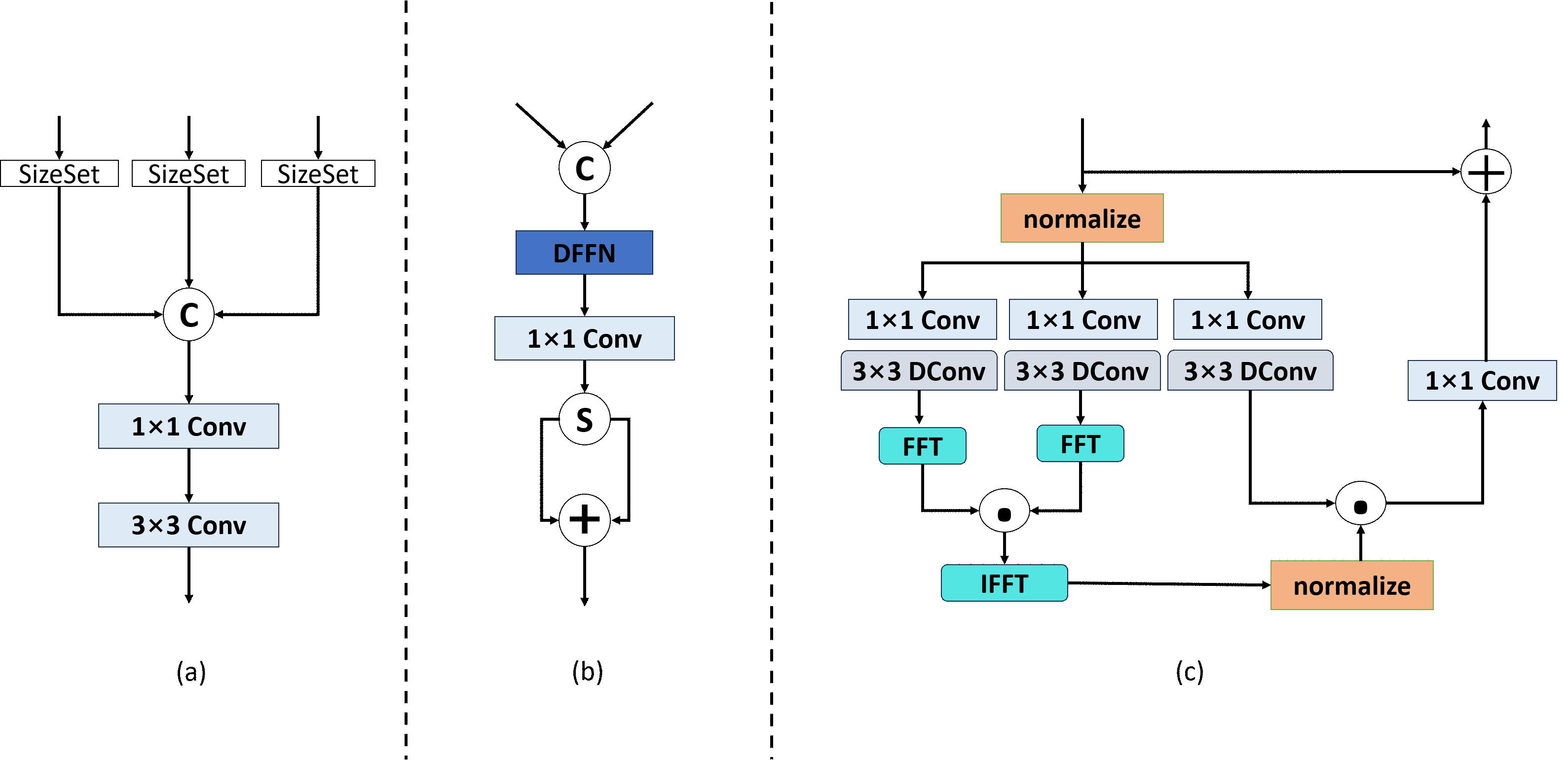}
\caption{The Structure of Decoder Block. (a) AFF, (b) Fuse, (c) FSAS.} \label{fig3}
\end{figure}

The main components of the decoder block are the DFFN module and the frequency domain-based self-attention solver (FSAS). For the FSAS module, the conventional visual transformer usually computes 
$F_{q}$ , $F_{k}$ and $F_{v}$ by applying linear transformations $W_{q}$ , $W_{k}$ and $W_{v}$ to the input features $\text{X}$ and then introduces the expansion function to extract $\{q_{i}\}_{i=1}^{N}$ ,   and $\{k_{i}\}_{i=1}^{N}$ and 
$\{v_{i}\}_{i=1}^{N}$ applies the reshaping operation to them to obtain the query $\text{Q}$ , keyword $\text{K}$ and value $\text{V}$ and realizes the extraction of the feature maps based on the three through the scaled dot product attention mechanism. The computational process is shown in Eq. (6).
\begin{equation}
    \begin{gathered}
    Q=R(\{q_{i}\}_{i=1}^{N}),K=R(\{k_{i}\}_{i=1}^{N}),V=R(\{\nu_{i}\}_{i=1}^{N})\\
    V_{att}=soft\max(\frac{QK^{T}}{\sqrt{CH_{P}W_{p}}})\cdot V
    \end{gathered}
\end{equation}
where $\text{R}$ is the reshaping function, $H_{P}$ and $W_{P}$ represent the height and width of the grabbed patch block, respectively.
This method is highly complex and computationally intensive, and its limitations will become apparent once the number of patches captured becomes large. Therefore, in the FASA module, $\text{Q}$ , $\text{K}$ and $\text{V}$ are obtained by a set of 1×1 convolution and 3×3 depth convolution, $\text{Q}$ and $\text{K}$ are multiplied element by element by FFT, which replaces $QK^{T}$ in Eq. (4), and greatly reduces its computational amount. The result is then inverted and multiplied by to obtain the estimated feature map. This realizes the efficient estimation of the attention map by element-by-element product operation in the frequency domain instead of matrix multiplication in the spatial domain. The detailed FSAS structure is shown in Fig.~\ref{fig3}
\subsection{Asymmetric U-shaped network architecture}
The asymmetric U-net structure is reflected in the asymmetry between the encoder module and the decoder module. As described in the previous subsection, the Decoder module contains the DFFN module and the FASA module, while only the DFFN module is used in the Encoder module. This is due to the fact that the Transformer module is more suitable for the decoder to perform long-range dependency capturing, thus generating higher quality results. While the encoder module extracts more shallow features, which usually contain blurring effects, if FSAS is applied to the encoder module instead, it will confuse the clear features with the blurred features, which is not favorable for image processing. Therefore the network adopts an asymmetric structure to achieve better deblurring effect. The encoder part follows Eq. (7) and the decoder part follows Eq. (8).
\begin{equation}
\hat{X}_{n}=DFFN(F_{EB}(X_{n},(\hat{X}_{n-1})^{\downarrow}))
\end{equation}
where $\hat{X}_{n}$ represents the output of the $n_{th}$ layer encoder, $F_{DB}(X_{n}^{'},(\hat{X}_{n+1}^{'})^{\uparrow}$ is the fusion of $X_{n}^{\prime}$ and $(\hat{X}_{n-1})^{\downarrow}$ with the corresponding encoder fusion, $X_{n}$ is the input of the $n_{th}$ layer encoder, and $(\hat{X}_{n-1})^{\downarrow}$ is the result of down-sampling the output of the $(n-1)_{th}$ layer encoder.
\begin{equation}
\hat{X}_{n}^{'}=DFFN(FSAS(F_{DB}(X_{n}^{'},(\hat{X}_{n+1}^{'})^{\uparrow})))
\end{equation}
where $\hat{X}_{n}^{'}$ denotes the output of the $n_{th}$ layer encoder, $F_{DB}({X}_{n}^{'},(\hat{X_{n+1}^{'}})^{\uparrow}$ is the fusion of ${X}_{n}^{'}$ and  $(\hat{X}_{n+1}^{'})^{\uparrow}$ in the corresponding encoder fusion mode, ${X}_{n}^{'}$ is the input of the $n_{th}$ layer encoder, and $(\hat{X}_{n+1}^{'})^{\uparrow}$ is the result of up-sampling the output of the $(n+1)_{th}$ layer decoder.
\section{Experiments}
\subsection{Dataset and implementation details}
In this paper, the GoPro~\cite{ref_lncs3} training dataset is used for training, which contains 2103 pairs of blurred clear control images.  The validation set used for training is 460 pairs of randomly divided images. The test set consists of 1111 pairs of blurry-clear images from GoPro and 48 pairs of blurry-clear images from Kohler~\cite{ref_lncs15} test dataset. 
In each training iteration, 4 images were randomly selected to be used as input after a random cropping operation to obtain an image of size 256×256. The learning rate was initially set to $10^{-4}$ and reduced by 0.5 times every 500 epochs. The model was trained on GoPro training set for 1200 iterations to get the final result. In addition to applying the model obtained from the GoPro training to the delineated GoPro test set, the same model was also tested by applying it directly to the Kohler test set as a way to check the generalization ability of the model. The graphics card used for the experiments was NVIDIA RTX 4090Ti.
\subsection{Performance comparison}
\begin{figure}
\includegraphics[width=\textwidth]{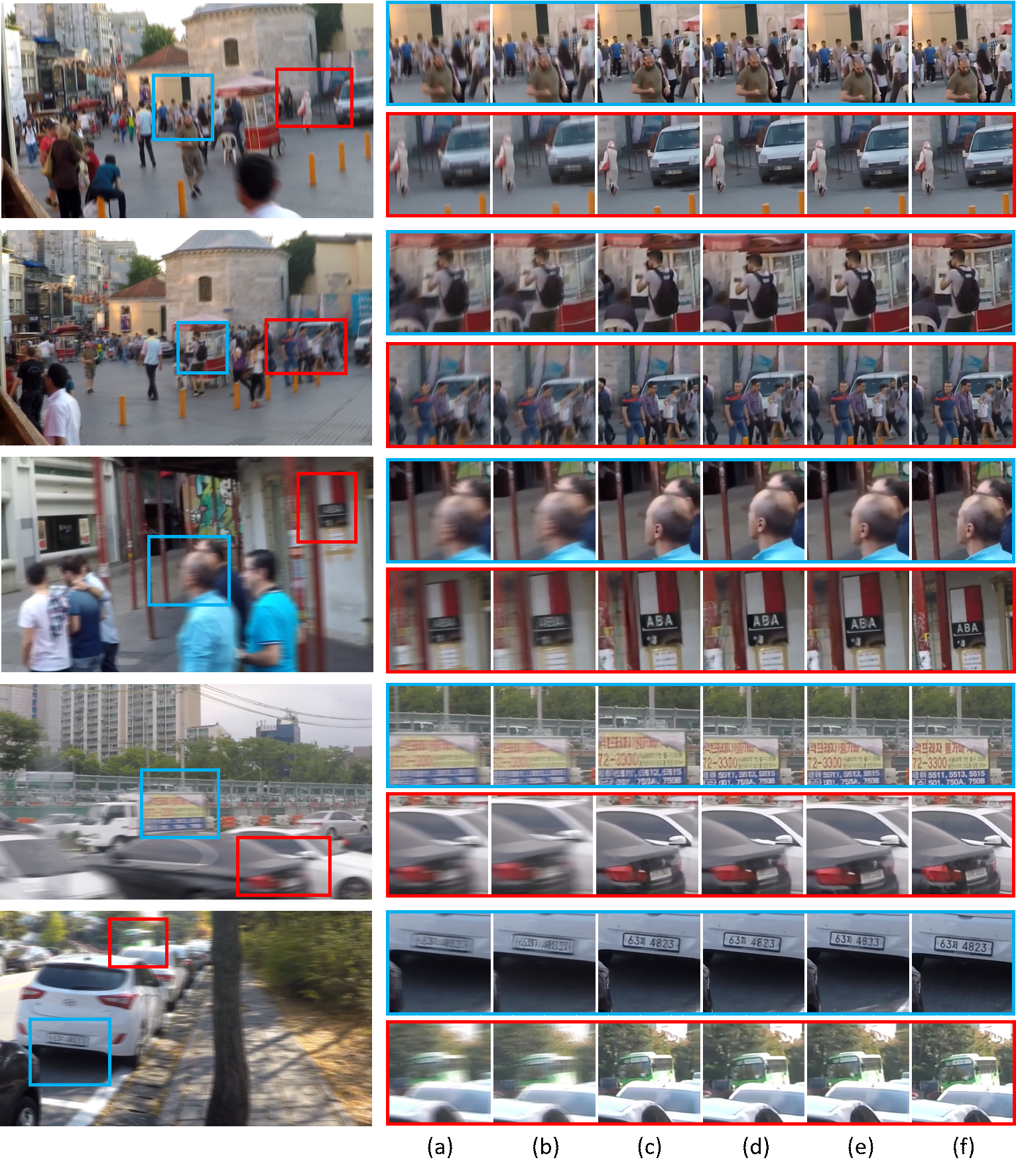}
\caption{Comparison of visualization on the GoPro. For clarity, the magnified parts of the resultant images are displayed. (a) blur, (b) DeblurGAN, (c) PSS-NSC, (d) Deepdeblur, (e) proposed method, (f) ground truth} \label{fig4}
\end{figure}

ASMA-UNet is compared with ~\cite{ref_lncs3,ref_lncs4,ref_lncs7,ref_lncs13,ref_lncs14,ref_lncs15,ref_lncs16,ref_lncs18}. The test results on the GoPro dataset are shown in Table~\ref{tab1}
\begin{table}
    \centering
    \caption{The test results on GoPro.}\label{tab1}
    \begin{tabularx}{\textwidth}{|X|X|X|X|}
    \hline
    Model &  PSNR & SSIM & Runtime\\
    \hline
    DeepDeblur~\cite{ref_lncs3} &  29.23 & 0.91 & 4.33s\\
    FCL-GAN~\cite{ref_lncs15} &  24.84 & 0.77 & \underline{0.01s}\\
    DeblurGANv2~\cite{ref_lncs14} &  29.55 & 0.93 & 0.35s\\
    DeblurGANv2~\cite{ref_lncs4} &  28.70 & 0.86 & 0.85s\\
    SRN~\cite{ref_lncs16} &  30.26 & 0.93 & 1.87s\\
    SVRNN~\cite{ref_lncs18} &  29.19 & 0.93 & 1.4s\\
    SVDN~\cite{ref_lncs13} &  29.81 & 0.93 & 0.01s\\
    PSS-NSC~\cite{ref_lncs7} &  \underline{30.92} & 0.94 & 0.99s\\
    \textbf{ASMA-UNet} &  \textbf{30.56} & \underline{\textbf{0.94}} & \textbf{0.05}\\
    \hline
    \end{tabularx}
\end{table}
the combined ability of ASMA-UNet proposed in this paper outperforms the rest of the deblurring methods. The accuracy is enhanced by the incorporation of the self-attention mechanism, the model makes excellent progress in capturing long-range dependency relations. The ASMA-UNet achieves an average PSNR of 30.55dB on the test set, which is 5.72dB, 1.33dB, and 1.01dB higher than the average PSNR values of the three adversarial network architecture-based models, namely FCL-GAN, DeblurGAN, and DeblurGANv2, and the average SSIM is also significantly improved compared to that of DeblurGAN and FCL-GAN, which are also significantly improved by 0.17 and 0.08 respectively. Compared to single network layer-based architectures such as SVRNN, SRN, and SVDN, the proposed method achieves average PSNR improvements of 1.37 dB, 0.30 dB, and 0.75 dB, respectively. These three methods benefit from a single network architecture and have certain advantages in processing speed over other methods. For instance, compared to deep convolutional methods like DeepDeblur, the average processing speed is improved by nearly 4s, with SVDN achieving an average runtime of just 0.01 seconds, which is 4.32 seconds faster than DeepDeblur. However, our method, which benefits from the introduced frequency-domain computation approach, achieves an average runtime of 0.05s. This is only 0.04 seconds slower than SVDN, while offering speed improvements of 1.83 seconds and 1.35 seconds over SRN and SVRNN, respectively. The PSS-NSC method, which is also based on the multi-scale processing method, has a slightly higher average PSNR of 0.36 dB, but its average processing time is nearly 1s more than that of the proposed method. In summary, ASMA-UNet achieves the goal of reducing the running time while guaranteeing the accuracy. The comparison results of representative methods from three classic networks—GAN, single-scale convolution, and multi-scale convolution—are shown in Fig~\ref{fig4}

It can be seen that compared with DeblurGAN and DeepDeblur and PSS-NSC methods, the method proposed in this paper has a better performance in visual aspect. From the results, the processing effect of this paper's method is obviously less artifacts and clearer edges than DeBlurGAN and DeepDeblur, which basically recovers the image contour and related information.
\FloatBarrier
\begin{figure}
\includegraphics[width=\textwidth]{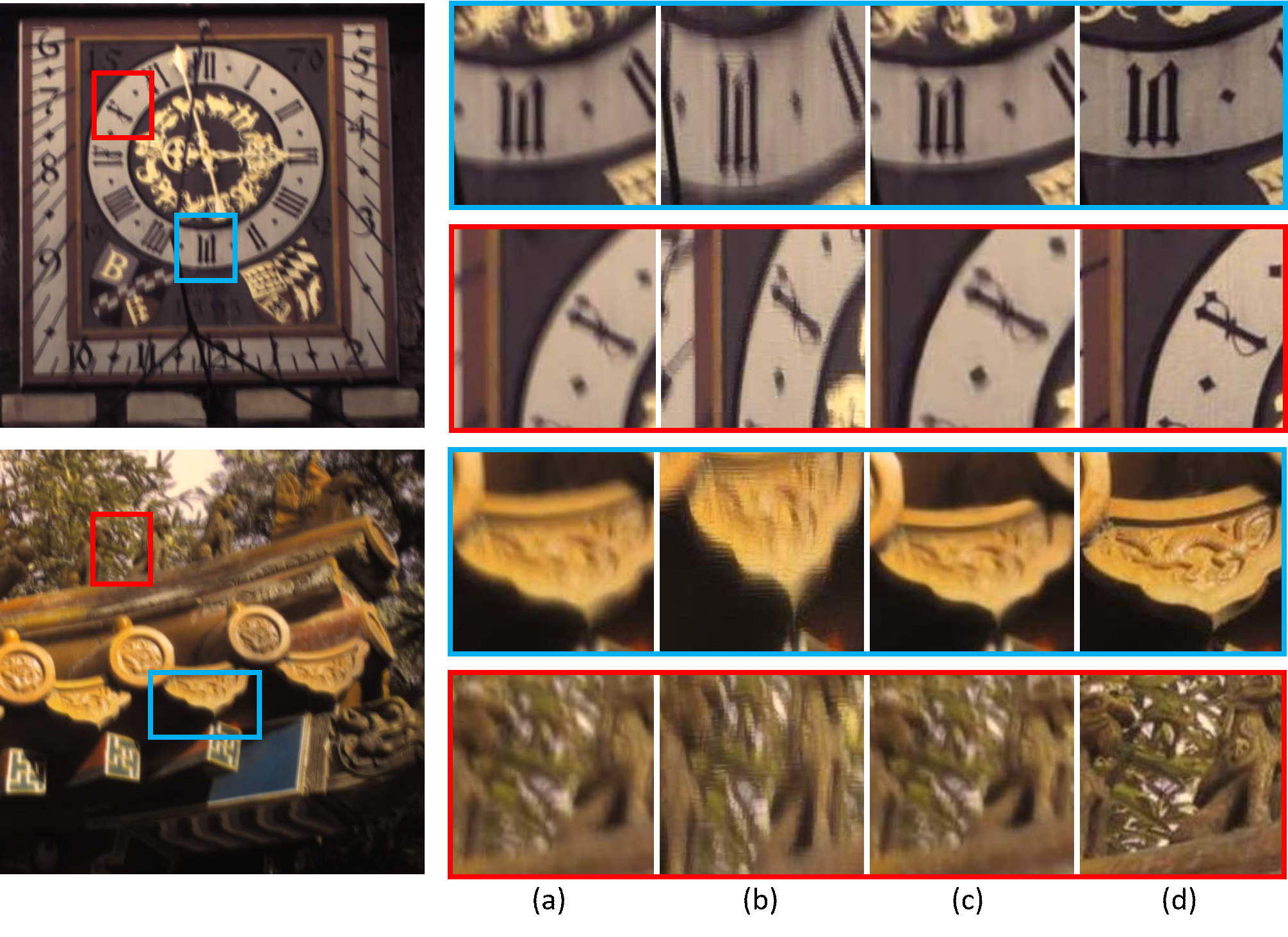}
\caption{The visual results directly applied to the Kohler. For clarity, the magnified parts of the resultant images are displayed. (a) blur, (b) DeblurGAN, (c) proposed method, (d) ground truth} \label{fig5}
\end{figure}
In terms of the performance of recovering text information in the distant view, the method in this paper has better processing effect compared with PSS-NSC. ASMA-UNet and DeblurGAN trained on the GoPro dataset were directly tested on the Kohler dataset. The results indicate that ASMA-UNet produces clearer edges and fewer artifacts and ripple noise compared to DeblurGAN. This demonstrates that the method proposed in this paper has stronger and more stable generalization capabilities. The visual results are shown in Fig~\ref{fig5}
\subsection{Ablation Experiment}
\begin{figure}
\includegraphics[width=\textwidth]{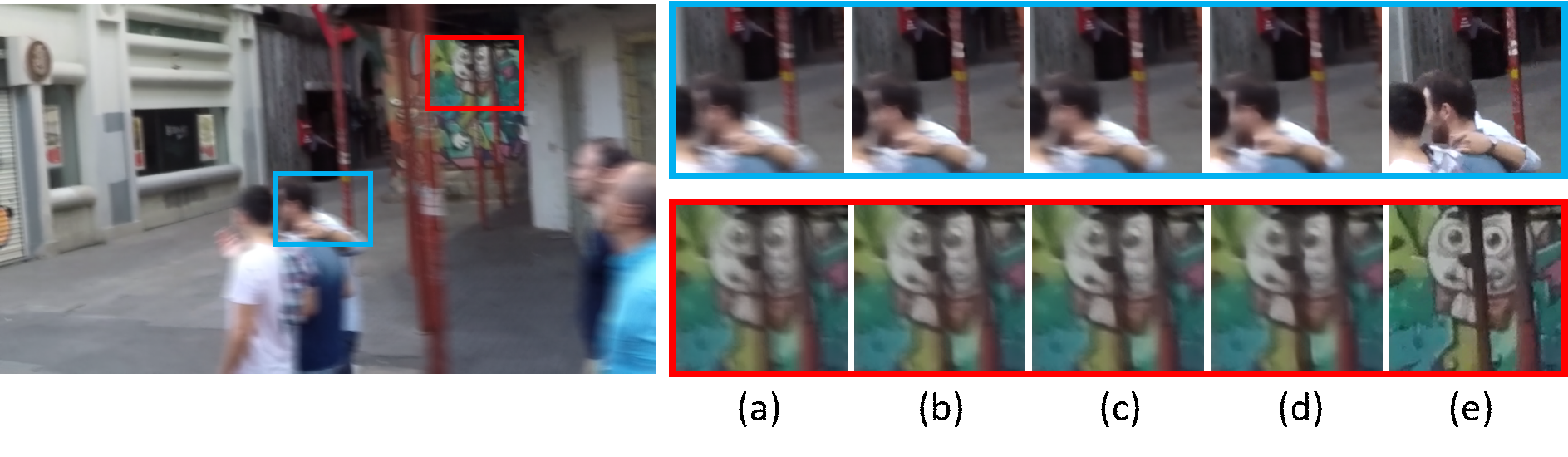}
\caption{The visual results. (a) AFF module removed, (b)a symmetric architecture, (c) original method, (d) ground truth} \label{fig6}
\end{figure}

To thoroughly analyze the effectiveness of the Asymmetric Multi-scale Feature Fusion (AFF) and the Asymmetric Architecture(AA) proposed in this paper, ablation experiments were conducted. For these experiments, a reduced dataset was randomly selected from the GoPro training set, consisting of 480 pairs of blurred and sharp images, along with a test set of 200 image pairs. Each component was evaluated over 150 iterations on the reduced dataset, and the comparative results are shown in Table 2. Firstly, a symmetric architecture was tested by adding the same DFFN modules to the encoder as in the decoder, forming a symmetric network structure. Secondly, the AFF module was removed so that each decoder layer's input was formed solely by upsampling the output from the preceding layer. The results are shown in the following Table~\ref{tab2}
\begin{table}
    \centering
    \caption{Results of the ablation experiments.}\label{tab2}
    \begin{tabularx}\textwidth{|X X|X|X|X|}
    \hline
    AA & AFF & PSNR & SSIM & Runtime\\
    \hline
      &  \textbf{\checkmark} & 27.10 & 0.86 & 0.06s\\
    \textbf{\checkmark} &  & 26.93 & 0.86 & 0.05s\\
    \textbf{\checkmark} &  \textbf{\checkmark} & \textbf{27.19} & 0.86 & 0.05s\\
    \hline
    \end{tabularx}
\end{table}
It can be observed that when the network architecture is symmetric, the PSNR decreases by 0.09 dB compared to the original network model, and the runtime increases by 0.01 seconds. This is directly related to the introduction of the DFFN in the encoder module, which not only causes a decline in experimental results but also leads to an increase in runtime. When the AFF is removed, the PSNR decreases by 0.26 dB compared to the original model, confirming the importance of the multi-scale fusion module in the network architecture for processing effectiveness. The visual results are shown in Fig~\ref{fig6}, demonstrating that the original model achieves the best visual performance.
\FloatBarrier
\section{Conclusion}
As described above, an asymmetric multiple scales U-net based on self-attention, incorporating a self-attention mechanism is proposed for deblurring tasks. Compared to methods that are purely based on convolutional neural networks or that directly integrate self-attention mechanisms, AMSA U-Net strikes a superior balance between accuracy and speed which can expand the receptive field of the model while enhancing computational efficiency, thus enabling effective deblurring. The decoder takes the multi-scale feature fusion result as input, processes it through a feedforward network that incorporates self-attention mechanisms, and produces an output. This enables the network to better capture feature information at different scales. Comparison results demonstrate that the proposed method significantly outperforms eight excellent methods.
\begin{credits}
\subsubsection{\discintname}
The authors have no competing interests. 
\end{credits}
%
%
%
%

\end{document}